\documentclass[a4paper,twoside]{article}

\usepackage{epsfig}
\usepackage{subcaption}
\usepackage{calc}
\usepackage{amssymb}
\usepackage{amstext}
\usepackage{amsmath}
\usepackage{amsthm}
\usepackage{multicol}
\usepackage{pslatex}
\usepackage{apalike}
\usepackage[bottom]{footmisc}

\usepackage{algorithm}
\usepackage{algorithmicx}
\usepackage{algpseudocode}

\usepackage{SCITEPRESS}     % Please add other packages that you may need BEFORE the SCITEPRESS.sty package.

\newcommand{\refeq}[1]{Eq.~\eqref{#1}}

\newcommand{\dimsym}{d}

\newcommand{\RN}{\mathbb{R}}
    
\DeclareMathOperator*{\loss}{{\ell}}

\newcommand{\mat}[1]{\mathbf{#1}}
\newcommand{\set}[1]{\mathcal{#1}}
\newcommand{\pnorm}[1]{\lVert{#1}\rVert}

\newcommand{\classifier}{\ensuremath{h}}
\newcommand{\x}{\ensuremath{\vec{x}}}

\newcommand{\xcf}{\ensuremath{\vec{x}_{\text{cf}}}}   
\newcommand{\deltacf}{\ensuremath{\vec{\delta}_{\text{cf}}}}    
\newcommand{\ycf}{\ensuremath{y_{\text{cf}}}}

\DeclareMathOperator{\sign}{sign}
\newcommand{\myCF}[3]{\ensuremath{\text{CF}_{#3}(#1, #2)}}

\begin{document}

\title{``How to make them stay?''-Diverse Counterfactual Explanations of Employee Attrition}

\author{\authorname{Andr\'e Artelt\sup{1,2}\orcidAuthor{0000-0002-2426-3126} and Andreas Gregoriades\sup{3}\orcidAuthor{0000-0002-7422-1514}}
\affiliation{\sup{1}Faculty of Technology, Bielefeld University, Bielefeld, Germany}
\affiliation{\sup{1}Department of Electrical and Computer Engineering, University of Cyprus, Nicosia, Cyprus}
\affiliation{\sup{3}Department of Management, Entrepreneurship and Digital Business, Cyprus University of Technology, Limassol, Cyprus}
\email{aartelt@techfak.uni-bielefeld.de, andreas.gregoriades@cut.ac.cy}
}


\abstract{
Employee attrition is an important and complex problem that can directly affect an organisation’s competitiveness and performance. Explaining the reasons why employees leave an organisation is a key human resource management challenge due to the high costs and time required to attract and keep talented employees. Businesses therefore aim to increase employee retention rates to minimise their costs and maximise their performance. Machine learning (ML) has been applied in various aspects of human resource management including attrition prediction to provide businesses with insights on proactive measures on how to prevent talented employees from quitting. Among these ML methods, the best performance has been reported by ensemble or deep neural networks, which by nature constitute black box techniques and thus cannot be easily interpreted. To enable the understanding of these models’ reasoning several explainability frameworks have been proposed to either explain individual cases using local interpretation approaches or provide global explanations describing the overall logic of the predictive model. Counterfactual explanation methods have attracted considerable attention in recent years since they can be used to explain and recommend actions to be performed to obtain the desired outcome. However current counterfactual explanations methods focus on optimising the changes to be made on individual cases to achieve the desired outcome.  In the attrition problem it is important to be able to foresee what would be the effect of an organisation’s action to a group of employees where the goal is to prevent them from leaving the company.  Therefore, in this paper we propose the use of counterfactual explanations focusing on multiple attrition cases from historical data, to identify the optimum interventions that an organisation needs to make to its practices/policies to prevent or minimise attrition probability for these cases. The proposed technique is applied on an employee attrition dataset, used to train binary classifiers. Counterfactual explanations are generated based on multiple attrition cases, thus, providing recommendations to the human resource department on how to prevent attrition .
}

\onecolumn \maketitle \normalsize \setcounter{footnote}{0} \vfill

\section{\uppercase{Introduction}}
\label{sec:introduction}

Employees constitute one of the most valuable assets in any organization. Therefore, the optimum way to manage this resource significantly improves organisational performance and competitiveness while also assists in obtaining organisation’s objectives. Human Resource (HR) management departments therefore engage in activities that aim to unleash employees’ full potential and maximise their productivity. Typical HR activities include the process of selection and recruitment, performance management, employee well being and satisfaction, training and development.
Due to the excessive cost associated with the selection, recruitment and training of employees before they become productive, HR management is constantly striving to keep their employees satisfied since the retention of talented employees is crucial to any company’s success. Retention involves the systematic effort of creating a working environment that satisfies employees’ needs by implementing appropriate policies and practices. Identifying optimum policies is a key problem in HR management and is the focus of this work. To address this issue, it is essential to understand what causes employees’ attrition that refers to the situation when an employee leaves the business either voluntarily or involuntarily. In the former case an employee makes a personal decision to leave the company and in the latter an employee is forced to resign due to low performance or not desired skills. Voluntary turnover (attrition) tends to relate with more skilled and talented employees and thus the loss for an organisation is significant in such cases due to loss of expertise which in certain cases might create operational issues.
Therefore, voluntary turnover can have direct and indirect effects to an organisation through increased hiring and training costs, reduced productivity, profits and employee morale. Therefore, employees’ intention to leave an organisation is a widely studied topic, with many studies investigating different factors that positively or negatively influence it. Such studies utilise questionnaires completed either by active employees to measure their turnover intention, or data from employees that provided their resignation notice explaining their motivations behind the decision to leave. Utilisation of turnover intention data however is considered the most feasible approach in attrition prediction since detailed resignation data is often unavailable due to privacy policies. Such studies test different hypotheses or provide new knowledge regarding the way different personal and organisational factors affect retention/attrition, mainly through statistical significance tests via regression and analysis of variance. However, such studies provide limited insights or recommendations on what organisations need to do to prevent attrition under different circumstances since their insights focus on how each factor affects attrition and not on what state these factors should be in collectively to prevent attrition. Such optimum states of attrition factors can be identified through interpretations of machine learning models trained on historical data and interrogated through explanation methods that provide a recommendation what to do in order to increase retention. 
Machine learning has been applied extensively in different prediction tasks including attrition~\cite{zhao2018employee} due to its improved prediction performance compared to traditional techniques. However, black box techniques tend to outperform the accuracy of interpretable methods such as decision trees. Thus, different explainability methods have been proposed to shade light into the logic of ensemble or deep neural network models, by identifying patterns in their reasoning~\cite{molnar2020interpretable,ExplainingBlackboxModelsSurvey,linardatos2020explainable}. Counterfactual explanations~\cite{CounterfactualWachter} (CE) are a popular technique for interpreting black box models that enable the identification of the optimum changes to be made to a model’s feature values to obtain the desired outcome.
The approach presented in this paper aims to identify the minimum consistent feature-value changes that reverse attrition prediction for a group of employees rather than individual cases and thus provide valuable recommendations to management on what to focus on maximising retention.  The identification of recommendations based on a group of cases (employee attrition) is what differentiates the method proposed in this paper from other studies that apply counterfactual explanations to the employee attrition problem. The data that we utilise to demonstrate the proposed approach is the IBM dataset~\cite{IbmHrAnalyticsEmployee} that is publicly available.

\section{\uppercase{Literature review}}
\label{sec:literaturereview}

Employee attrition prediction using ML has gained significant attention in recent years, with scholars utilising a wide range of ML techniques on work-related factors to predict employee turnover intention~\cite{zhao2018employee}. Indicatively the following ML techniques have been applied to the employee attrition prediction problem: Support vector machines, Decision trees, Random Forest, XGBoost, Logistic Regression, Näıve Bayes, Adaptive Boosting, K-nearest neighbors, Artificial neural network, and Light Gradient Boosting. To the best of our knowledge the best performance is reported with the SVM and LGB.
 Prevalent features used to train such models relate to individual and organisational factors such as number of promotions, salary, last evaluation, time spent in company, working conditions, working hours, employee-related factors such as age, gender, job satisfaction, work life balance, emotional exhaustion, growth potential, and marital status. These factors have been identified in previous works~\cite{kang2021predictors,le2022turnover} that examined how individual and organisational factors affect employee attrition. The effect of the combination of these factors varies.   For instance,  factors such as time in service showed that employees who have worked for longer at a given organization are less likely to leave. While longer time before employees obtain organizational tenure positively mitigates to employees’ intentions to leave. Sex and education have also been identified as influencing factors however there is no consensus on how each factor affects attrition. Organizational factors on the other hand are factors controlled by the organisation and thus could be altered to improve retention. Such factors are job satisfaction that is strongly and significantly related to employees’ intentions to leave, with employees who feel higher personal accomplishment in their job being less likely to leave their organization. Overloading on the other hand results in emotional, physical, and mental exhaustion and thus increases employees’ intentions to leave. Employees who think of their work as meaningful are less likely to leave. Finally, employees who understand the relationship of their job to an organizations' goals are also less likely to leave their organization. Since they consider themselves as part of a team with a common objective.
Several organisational policies have been reported to improve retention such as salary, family-friendly policies, training, skill development, and diversity management. Specifically, rewards such as salary, increase employees’ motivation and satisfaction which makes them less likely to leave their organizations in contrast to employees that are not satisfied with their salary. In addition, employees who are not rewarded for their good performance are more likely to leave the organization. On the other hand, balancing work and family responsibilities reduce employees’ turnover. While training and development opportunities are found to mitigate turnover but can also increase turnover since employees become  more attractive by competitors. Working environment on the other hand, such as good relationships with co-workers and fairness in organisational procedures reduce intention to leave, while organizations that support creativity and innovation are significantly related to lower turnover intentions due to the associated sense of meaning and pleasure from the work done. While, employees who feel supported by their organization are less likely to leave the organization since they consider that the organisation takes care of them. Finally, a meta-analytic review of voluntary attrition~\cite{haldorai2019factors} found that the strongest predictors of employee turnover are, age, pay, and job satisfaction. Other studies  highlight that  ``anti-social'' working hours, work life conflict, emotional exhaustion, work overload,  working environment,  career progression and community fit strongly influencing turnover intention.

\section{\uppercase{Counterfactual explanations}}
\label{sec:counterfactuals}

In predictive modelling it is important to have good accuracy of what is predicted but also high interpretability. The former is essential to be able to make actionable decisions while the latter increases the confidence \& trust in the predictions and understanding of the model’s logic. However, most models that produce high predictive accuracy are also less interpretable and vice versa. Several (explanation) methods have been developed to increase the interpretability of black box models~\cite{molnar2020interpretable}, however, no practical technology has yet emerged for explainable AI~\cite{ExplainingBlackboxModelsSurvey,linardatos2020explainable}. This is due to the complexity of the matter with explanations having to be both statistically sound and comprehensible by stakeholders.
Many approaches to the explainability problem focus on the global logic of a black box model through an associated interpretable classifier such as decision tree that mimics the behavior of the black box model. These methods are model dependent. A different branch of work on model-independent (agnostic) techniques for understanding black box models' behaviour is by using the classifiers' predictions to generate explanations. These are referred as post hoc  since the technique is  applied  after model training. Such model agnostic techniques are further categorised into local and global explanation methods. The local focusing on specific instances and global on the whole set of instances. A few recent methods that are model-agnostic, such as LIME~\cite{ribeiro2016should} obtain a local explanation for a decision outcome by learning an interpretable model from querying randomly perturbed versions of a given instance on a black box model. Such post-hoc methods for extracting local explanations from each model prediction have been attracting much attention.
Counterfactual explanation~\cite{CounterfactualWachter,verma2020counterfactual} is a post-hoc local explanation method that show why the undesirable predictions emerged and what needs to be changed in the input to obtain the desired results. Thus a counterfactual explanation describes a causal situation in the form: \textit{If X had occurred, Y would have occurred}. Counterfactual explanations can be used to explain predictions of individual instances and determine required input to obtain opposite results. 

To generate counterfactuals researchers use optimization-based approaches. Therefore, once you have a trained classifier the goal of the optimiser is to find a counterfactual  with the shortest distance to a case with the desired output (non attrition). Such counterfactual can be obtained by solving an optimization problem.

Various local explanation methods however have been criticized for not being robust~\cite{artelt2021evaluating,hancox2020robustness,mishra2021survey} or that they might fail to explain the global behavior of complex models ~\cite{slack2021counterfactual}. Thus, such methods have limited applications in problems such as employee attrition that require the identification of the optimum changes to be made by the HR department to prevent attrition. For such a problem, multiple instances of employees that quit their jobs need to be considered.  
To address the above issue, several methods for tackling the multiple local explanations problem have been proposed. For instance,  AReS~\cite{rawal2020beyond},which is a global summary of  actions expressed in rules and MAME (Rama-murthy et al., 2020) and GIME~\cite{gao2021learning} that are global summaries of general local explanation methods (e.g., LIME~\cite{ribeiro2016should}).
Recently, another approach~\cite{kanamori2022counterfactual} for simultaneously computing counterfactual explanations for different groups of individuals was proposed that is similar to the approach presented in this paper.

\section{\uppercase{Methodology}}
\label{sec:methodology}

We propose a counterfactual explanation methodology for computing a recommendation (i.e. explanation) of what the HR department should do to increase retention of employees. Essentially the goal is to find the changes the HR departments must do to their policy such that many (if not all) of the employees that initially intended to leave the company change their minds. For this purpose we assume that we have a set of employees where an attrition classifier $\classifier: \RN^\dimsym \to \{-1,1\}$ predicts that they 
 quit their jobs. We denote this set of employees as $\set{D}$ and assume that each employee is represented by some real-valued feature vector $\x_i\in\RN^\dimsym$.

\subsection{A single explanation}\label{sec:methodology:singlecf}
We are looking for the changes the HR department must make  $\deltacf\in\RN^\dimsym$ which, if applied to the employees $\x_i$, change the prediction of the attrition classifier $\classifier(\cdot)$ from attrition to retention. We call 
 such a $\deltacf$ a counterfactual explanation of attrition. We phrase this as the following optimization problem -- since we are looking for a ``simple'' (i.e. cost sensitive) recommendation of changes $\deltacf$, we aim to minimize the number and magnitude of changes in $\deltacf$ by using the $1$-norm\footnote{The $1$-norm treats all features as equally important and costly -- this can be changed by using a weighted $1$-norm instead.}:
\begin{equation}\label{eq:setconsistentcf:relaxed}
    \underset{\deltacf\,\in\,\RN^\dimsym}{\min}\; \pnorm{\deltacf}_1 + C\cdot\sum_{\x_i\in\set{D}}\loss\left(\classifier(\x_i + \deltacf)\right)
\end{equation}
where $\loss(\cdot)$ denotes a suitable loss function penalizing attrition predictions (i.e. penalty if the change $\deltacf$ does not flip the output  of the model for an employee). Suitable loss functions might be mean-squared error or cross-entropy loss.
The regularization strength $C>0$ allows us to balance between the two objectives of having a ``simple''  recommendation of actions and a recommendation that works for as many employees as possible.

While our formalization~\refeq{eq:setconsistentcf:relaxed} is completely model-agnostic -- i.e. it can be applied to any attrition classifier $\classifier(\cdot)$, it might not be the most efficient formalization for every situation. In particular if we have knowledge about $\classifier(\cdot)$ that could be exploited for a more efficient (e.g. faster) computation of the recommendation $\deltacf$,  this can used. In the following, we investigate the special case of linear classifiers:
In case of a linear classifier $\classifier(\x) = \sign\left(\vec{w}^\top\x + b\right)$ such as linear-SVM, logistic regression, etc., we can rewrite~\refeq{eq:setconsistentcf:relaxed} as the following linear program (LP):
\begin{equation}\label{eq:setconsistentcf:linearclassifier:relaxed}
\begin{split}
    &\underset{\deltacf\,\in\,\RN^\dimsym}{\min}\; \pnorm{\deltacf}_1 + C\cdot\sum_i\xi_i \\
    &\text{s.t. } \ycf\cdot\left(\vec{w}^\top\deltacf + \vec{w}^\top\x_i + b\right) \geq 0 - \xi \quad \forall\,\x_i\in\set{D}\\
    & \xi \geq 0 \quad\forall\,i
\end{split}
\end{equation}
where $C>0$ is again a hyperparemeter that allows us to balance between the two objectives. In practice it might be necessary to try different values for $C$ in order to find a practically satisfying/useful recommendation $\deltacf$.
Note that linear programs (LP) are special instances of convex optimization problems that can be solved very efficiently~\cite{boyd2004convex}.

\subsection{A set of diverse explanations}\label{sec:methodology:diversecf}
As already noted by~\cite{kanamori2022counterfactual}, there often might not exist a single change $\deltacf$ that is applicable to all employees in $\set{D}$. Furthermore, there might also exist several different changes $\deltacf$ that work equally well (and maybe also work for different subgroups of employees), we therefore propose an extension to our formalization from Section~\ref{sec:methodology:singlecf} that computes not a single recommendation of changes $\deltacf$ but a set of different \& diverse changes $\deltacf$ so that the decision makers are provided with a list of possible actions on how to increase retention rate. Therefore, the HR manager can choose the action that is more suitable to his/her case.

We propose an iterative method that computes a set of highly diverse changes $\deltacf$. We defined diversity by means of the number of overlapping features. For instance, highly diverse explanations should use \& change completely different features. For this purpose, we need a mechanism for excluding already used features from future changes $\deltacf$. We can modify our previous optimization problems not to use any black-listed features $\set{F}$ -- i.e. features already used by previous recommendations of changes -- by introducing a diagonal matrix $\mat{M}\in\RN^{\dimsym\times\dimsym}$ and replacing all occurrences of $\deltacf$ by
\begin{equation}\label{eq:ignoreblacklistedfeatures}
    \mat{M}\deltacf
\end{equation}
The diagonal matrix $\mat{M}$ is then defined as follows:
\begin{equation}
    (\mat{M})_{i,i} =
    \begin{cases}
    0  & \quad \text{if } i\in\set{F}\\
    1  & \quad \text{otherwise}
  \end{cases}
\end{equation}
The effect of~\refeq{eq:ignoreblacklistedfeatures} is that it sets all changes in black-listed features $\set{F}$ to zero -- i.e. it basically removes forbidden changes and the solver is required to find other changes $\deltacf$ that yield a feasible solution.
Note that~\refeq{eq:ignoreblacklistedfeatures} does not change the computational complexity of the original optimization problems~\refeq{eq:setconsistentcf:relaxed} and~\refeq{eq:setconsistentcf:linearclassifier:relaxed}.
However, it can happen that for a set of black-listed features $\set{F}$ no feasible solution exists -- i.e. no allowed changes will change the attrition prediction.

We introduce the following notation: $\myCF{\set{D}}{\set{F}}{\classifier}$ computes the solution to one of the previous optimization problems (i.e. depending on $\classifier(\cdot)$ either~\refeq{eq:setconsistentcf:relaxed} or~\refeq{eq:setconsistentcf:linearclassifier:relaxed}) under the additional constraint that no black-listed features $\set{F}$ must be used in the final solution.
The complete procedure is described in pseudo-code in Algorithm~\ref{algo:diverse_cf}.
\begin{algorithm}%[t]
\caption{Computation of Diverse Counterfactual Explanations of Attrition}\label{algo:diverse_cf}
\textbf{Input:} Set of employees $\set{D}$ where attrition is predicted; $k\geq 1$: number of diverse counterfactual explanations of attrition; attrition classifier $\classifier(\cdot)$ \\
\textbf{Output:} Set of diverse counterfactuals $\set{R}=\{\xcf^i\}$
\begin{algorithmic}[1]
 \State $\set{F} = \{\}$  \Comment{Initialize set of black-listed features}
 \State $\set{R} = \{\}$  \Comment{Initialize set of diverse counterfactuals}
 \For{$i=1,\dots,k$} \Comment{Compute $k$ diverse counterfactuals}
    \State $\deltacf^i = \myCF{\set{D}}{\set{F}}{\classifier}$  \Comment{Compute next recommendation of changes}
    \State $\set{R} = \set{R} \cup \{\deltacf^i\}$
    \State $\set{F} = \set{F} \cup \{j \mid (\deltacf^i)_j \neq 0\}$  \Comment{Update set of black-listed features}
 \EndFor
\end{algorithmic}
\end{algorithm}

\subsection{Related work}
The authors of~\cite{kanamori2022counterfactual} deal with a very similar problem, which they call ``group-wise counterfactual explanations''. However, the authors do not try to find a single explanation applicable to as many as possible employees like we do in Section~\ref{sec:methodology:singlecf}, but instead propose an algorithm/heuristic called ``counterfactual explanation tree'' for partitioning the employees into groups for which a single applicable explanation (i.e. recommendation of changes $\deltacf$) is computed.
However, their proposed method is computationally expensive and the algorithm for building the counterfactual explanation tree is just a heuristic that comes without any formal guarantees. Furthermore, their algorithm can not be easily extended or customized -- e.g. adding constraints to the explanations or computing multiple explanations -- whereas our optimization as illustrated in  Section~\ref{sec:methodology:singlecf} can be extended and reused in many different ways as we do in Section~\ref{sec:methodology:diversecf} for computing a set of diverse changes $\deltacf$.

\section{\uppercase{Case study}}
\label{sec:experiments}
To illustrate the application of the proposed multi instance counterfactual method in the employee attrition problem the following data-set is utilised.

\subsection{Dataset}
The IBM human resource dataset~\cite{IbmHrAnalyticsEmployee} contains $35$ features for $1467$ unique employees. The dataset contains human resources properties such as age, education, gender, promotion, education and rate. The question we want to answer using this data is ``what does the HR department need to do to prevent employees from leaving the organisation''. To address this question all the attrition cases need to be considered and the optimum solution that will satisfy all these cases is identifying using the proposed method. The method utilise only features that can be changed such as salary, years before promotion, business travel, environmental conditions, etc. Additionally, for this example we utilise only numerical features since they are easier to manipulate and thus provide proof of concept. However categorical features can also be utilised after undergoing one-hot encoding or when converted to ordinal variables (e.g. education can be expressed as an ordinal variable with high values indicating education of PhD level).  Features that can not be changed such as age, gender etc are not utilised. However these features can be utilised when hiring employees, so as to target employees that are more likely to stay at the organisation. 
The method can also be used with features that are generated from combinations of existing features, for example a new feature ``job hopper'' that refers to people that change jobs regularly to increase their salary and thus may have a higher probability of leaving the company can also be generated and utilised in the analysis. This can be estimated from the number of companies the employees worked before divided by the total working years. 

\subsection{Setup}
We implement logistic regression, random forest and XGBoost as attrition classifiers using the IBM dataset. To account for the differences in  job roles and departments the analysis can be performed by focusing on cases refering to different departments such as "Research and development" or "Sales". This could give more specific recommendations that can relate better to the department in question.For this case study we focused in the Research and Development department and the features that are  utilized are:  \textit{EnvironmentSatisfaction},  \textit{JobInvolvement}, \textit{JobSatisfaction}, \textit{MonthlyIncome}, \textit{PercentSalaryHike}, \textit{YearsInCurrentRole}, \textit{YearsSinceLastPromotion}, \textit{YearsWithCurrManager}. These features have been known to affect attrition based on the literature, thus we aimed to identify how the HR department can alter the intention of employees by changing individual or combinations of these factors. The contribution is the intensity of the change and the combination of the features to achieve the desired outcome. All classifiers have been optimised using hyper-parameter tuning. Data is split into train and test set for finding the optimal hyper parameters. Furthermore, data is standardized and random under-sampling of the majority class is performed in order to avoid any biases in the classifier due to class imbalance.

\subsection{Results}
We observe that our method (Section~\ref{sec:methodology}) computes several reasonable recommendations (of different complexity) on how to reduce employee attrition and provided valuable insights on employee turnover which is consistent with the literature. Specifically, the following recommendations refer to turnover instances and indicate the features that must be manipulated and by how much to prevent employees from leaving the company:
\begin{enumerate}
    \item If employees would have had \textit{an increase in Percent Salary Hike of approx. $40  \%$}, attrition would be unlikely.
    \item If employees would have had \textit{approx. $5$ years less since their last promotion}, attrition would be unlikely.
    \item If employees would have had \textit{an increase in Percentage Salary Hike of approx. $20  \%$,  AND had in an increase in job satisfaction by approx. $50  \%$ }, attrition would be unlikely.

\end{enumerate}

These recommendations highlight the importance of salary, job satisfaction and promotion as key factors for preventing employee turnover, which abide with the theory or retention.  Specifically, the results from the example case study show that to achieve this goal the company needs to increase the salary  of selected  employees or change its promotion policy so as to motivate employees by promoting them earlier. However, given that the recommended salary increase is large, companies could apply the recommendation on talented employees that can not afford to loose. Similar actions are recommended by ~\cite{kanamori2022counterfactual}, thus highlighting the importance of salary (monthly income) as key factors for retention, while also provides confidence in our preliminary results. In addition, as highlighted in the literature the method also identifies that job satisfaction is a key property  for preventing attrition if combined with salary increase. Increase in job satisfaction can be realised through improved working conditions such as renovation of the working environment, bonding activities, organised events, complimentary food etc. 

In contrast to other emplainable ML methods such as SHAP ~\cite{lundberg2017unified} that provide global explanations  and specify the factors that influence the class variable (attrition), the proposed method  can go one step further by providing specific  recommendations of business policy changes to revert attrition of specific employees.
These  provide actionable business recommendations that in contrast to  SHAP provide specific policy changes that are optimised on multiple instances (for instance employees in specific department that is key to the competitiveness of the company and their skills cannot easily be replenished) and explain what the management needs to do to  prevent attrition of these employees.

\section{\uppercase{Conclusions}}
\label{sec:conclusion}
The method proposed in this work utilise counterfactual explanations but in contrast to other similar studies utilise multiple instances to generate recommendations that are actionable by businesses. This is important in problems such as employee attrition where specific HR actions on what needs to be done to prevent talented employee attrition should be based on multiple instances rather than individual employees that left the company. In the case of a single employee attrition prevention, the company needs only to consider the properties of an individual to derive actions on how to prevent him/her from quitting, however, in reality the company needs a general policy to reduce attrition for all turnover cases due to the negative effect that  attrition has on business performance and competitiveness. The method is applied on the IBM attrition dataset showing policy changes that are valid since they abide with the literature. Future work will focus on providing weights to the features that are manipulated so as for the method to provide recommendations ranked based on how  easy these can be implemented. For instance salary increase by 30 percent might be more difficult to realise than increase in company's environmental conditions. In addition the method will be extended to eliminate automatically infeasible solutions by enabling the user to provide range of feasible feature values.   Future work  will also focus on comparing the results of our method with similar work under different scenarios.

\section*{\uppercase{Acknowledgements}}

We gratefully acknowledge funding from the VW-Foundation for the project \textit{IMPACT} funded in the frame of the funding line \textit{AI and its Implications for Future Society}.

\bibliographystyle{apalike}
{\small
\bibliography{example}}

%\section*{\uppercase{Appendix}}
%
%TODO

\end{document}